\documentclass{article}

    \usepackage[preprint]{neurips_2022}

\usepackage[utf8]{inputenc} 
\usepackage[T1]{fontenc}    
\usepackage{hyperref}       
\usepackage{url}            
\usepackage{booktabs}       
\usepackage{amsfonts}       
\usepackage{nicefrac}       
\usepackage{microtype}      
\usepackage{xcolor}         
\usepackage{amsmath}
\usepackage{graphicx}

\title{Beta-VAE has 2 Behaviors: PCA or ICA?}

\author{
  Zhouzheng.~Li\\
  Department of Electrical \& Mechanical Engineering\\
  Beijing University of Chemical Technology\\
  Beijing, China \\
  \texttt{zhouzheng\_li@buct.edu.cn} \\
  \And
  Hao.~Liu \\
  Department of Mechanical \& Industrial Engineering\\
  University of Toronto \\
  Toronto, Ontario, Canada \\
  \texttt{liuhao.liu@mail.utoronto.ca} \\ 
}

\begin{document}

\maketitle

\begin{abstract}
  Beta-VAE is a very classical model for disentangled representation learning, the use of an expanding bottleneck that allow information into the decoder gradually is key to representation disentanglement as well as high-quality reconstruction. During recent experiments on such fascinating structure, we discovered that the total amount of latent variables can affect the representation learnt by the network: with very few latent variables, the network tend to learn the most important or principal variables, acting like a PCA; with very large numbers of latent variables, the variables tend to be more disentangled, and act like an ICA. Our assumption is that the competition between latent variables while trying to gain the most information bandwidth can lead to this phenomenon.
\end{abstract}

\section{Introduction}

\subsection{Beta-VAE}

Beta-VAE, proposed by \cite{Higgins2016betaVAELB}, is a type of variational autoencoder, which is a neural network that learns to compress and reconstruct data. It is called "beta" because it involves a regularization parameter, $\beta$, that controls the trade-off between the reconstruction accuracy and the information that is retained in the compressed representation of the data.

The goal of beta-VAE is to learn a compressed representation of the input data that captures the underlying factors of variation in the data. This is achieved by adding a penalty term to the loss function of the autoencoder, which encourages the compressed representation to be more disentangled, i.e., to have separate dimensions that capture different factors of variation.

In other words, beta-VAE tries to learn a representation that separates the different sources of variation in the data, such as shape, color, texture, etc., into distinct dimensions. This makes the compressed representation more interpretable and easier to manipulate for downstream tasks, such as generating new samples or editing existing ones.

Beta-VAE has been used in various applications, such as image generation, video analysis, and robotics, and has been shown to be effective in disentangling the factors of variation in the data.

\subsection{PCA}

PCA (Principal Component Analysis), \cite{Jolliffe2002PrincipalCA} is a statistical technique used for reducing the dimensionality of a dataset. It works by identifying the underlying structure of the data and finding a set of new, orthogonal (uncorrelated) variables, known as principal components, that capture the most significant variations in the data.

PCA is often used to preprocess data in machine learning and data analysis tasks to reduce the complexity of the data and make it easier to analyze. It can also be used for data compression, visualization, and noise reduction.

In PCA, the first principal component is the direction in the data that has the largest variance, and each subsequent principal component captures the maximum remaining variance in the data, subject to the constraint that it is orthogonal to the previous components. The principal components can be visualized as a set of axes that define a new coordinate system for the data.

PCA is based on linear algebra and involves calculating the eigenvectors and eigenvalues of the covariance matrix of the data. The eigenvectors represent the principal components, and the eigenvalues represent the amount of variance captured by each component.

PCA can be applied to various types of data, including numerical data, images, and text. It is a powerful tool for exploratory data analysis, dimensionality reduction, and feature extraction, and is widely used in many fields, including finance, engineering, and biology. The problem with PCA is that it only works with linear data.

\subsection{ICA}

ICA (Independent Component Analysis), \cite{Hyvrinen2001IndependentCA} is a statistical technique used for separating a multivariate signal into independent, non-Gaussian components. It is a type of blind source separation, which means that it can separate the components without any prior knowledge of their sources or the mixing process that created the signal.

The goal of ICA is to find a linear transformation that decorrelates the components of the input signal and maximizes their independence. It is often used in signal processing applications, such as audio and image processing, to extract meaningful signals from noisy or mixed data.

ICA assumes that the observed signals are a linear mixture of independent sources, which can be expressed as a matrix multiplication of the mixing matrix and the source signals. The mixing matrix represents the process that mixes the sources to produce the observed signals, and the goal of ICA is to estimate the sources from the observed signals by inverting the mixing matrix.

ICA is a computationally intensive technique and requires a large amount of data to estimate the mixing matrix accurately. It also assumes that the sources are non-Gaussian and statistically independent, which may not always hold true in real-world applications.

ICA has been applied to various fields, including speech processing, image analysis, and neuroscience. It is a powerful tool for separating signals and identifying meaningful patterns in complex data. Like PCA, ICA also works best with linear data.

\section{Experiments}

\subsection{Model setup}

The beta-VAE used in this paper is slightly different from that of \cite{Burgess2018UnderstandingDI}. While the overall loss is the same:

\begin{equation}
\mathcal{L}(\boldsymbol{x}, \boldsymbol{\hat{x}}, \boldsymbol{z}) = \underbrace{\frac{1}{2} ||\boldsymbol{x} - \boldsymbol{\hat{x}}||_2^2}_\text{reconstruction loss} + \underbrace{\beta D_{KL}(q(\boldsymbol{z}|\boldsymbol{x}) || p(\boldsymbol{z}))}_\text{regularization term}
\end{equation}

In our own version, the parameter $\beta$ is adjusted with a staircase and gets smaller exponentially. $\beta$ follows a schedule like this:

\begin{equation}
\beta = 0.917^{\beta_{init} + \lfloor \frac{iter}{100}\rfloor}
\end{equation}
where $\beta_{init}$ is the initial value we give beta and $iter$ is the current iteration number. 100 means for each 100 iterations the beta is shrunken once. $\beta_{init}$ is required so that the punishment is strong enough and that the bottleneck can remain very small (almost no information can pass) at the beginning. 0.917 means that the $\beta$ is approximately halved after 8 shrinks.

The model is constructed fully with fully connected layers with SeLU activation, trained with Adam optimizer and learning rate of 2e-3. We discovered that the learning rate needs to be bigger to quickly allow latent variables to seize the free information passage, so as to make sure the strongest latent variable always get the most resources. 

\subsection{Linear Dataset}

To compare with the PCA and ICA which only works on linear data, we designed a linear dataset with 4 inputs $y$ and 14 outputs $x$, 10000 $y$ is uniformly distributed between $[0,1]$, and $x$ is calculated with $y$ via a weight matrix:

\begin{equation}
X = Y\times W
\end{equation}
where $W$ is the weight matrix with shape $[4, 14]$.

\subsubsection{PCA and ICA Results}

\begin{figure}
  \centering
  \includegraphics[width=0.4\textwidth]{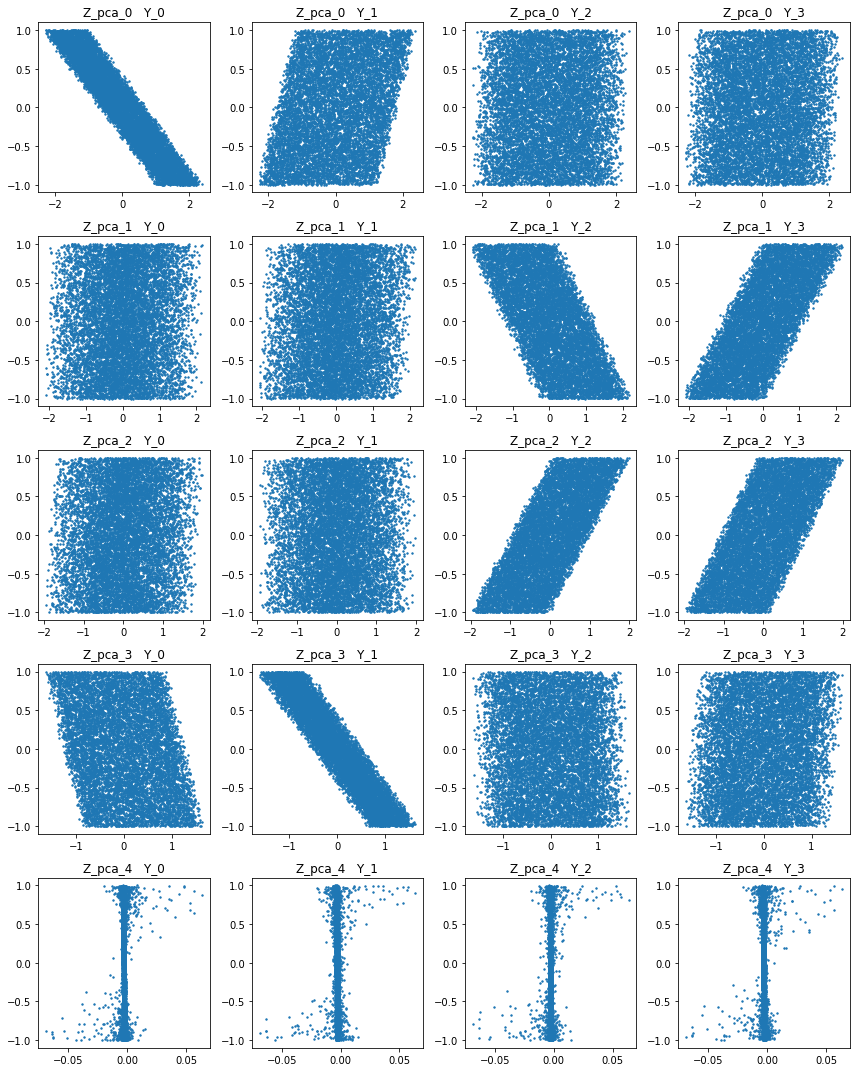}
  \caption{The components of PCA (5 dimensional) $Z_{pca0}$ to $Z_{pca4}$ with respect to the 4 input variables $Y_0$ to $Y_3$.}
  \label{fig:2}
\end{figure}

\begin{figure}
  \centering
  \includegraphics[width=0.4\textwidth]{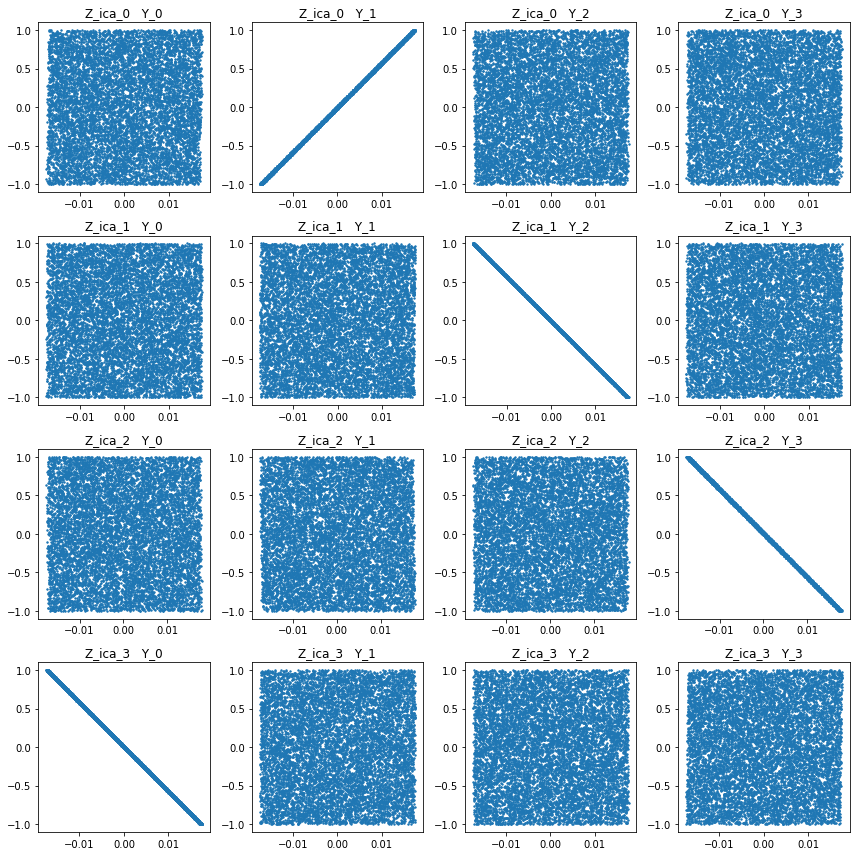}
  \caption{The components of ICA (4 dimensional) $Z_{ica0}$ to $Z_{ica4}$ with respect to the 4 input variables $Y_0$ to $Y_3$. The disentanglement is pretty obvious.}
  \label{fig:3}
\end{figure}

First, the PCA and ICA results for the linear data are presented. We give PCA 5 components (the 5th variable of PCA shouldn't learn anything useful since the system is only 4 dimensional), and 4 components for ICA (any number larger than 4 will cause the ICA to not converge). Then we compare the results of PCA and ICA with the label Y, as shown in Fig.\ref{fig:2} and Fig.\ref{fig:3}.

\subsubsection{Few Latent Variables Scenario}

\begin{figure}
  \centering
  \includegraphics[width=0.6\textwidth]{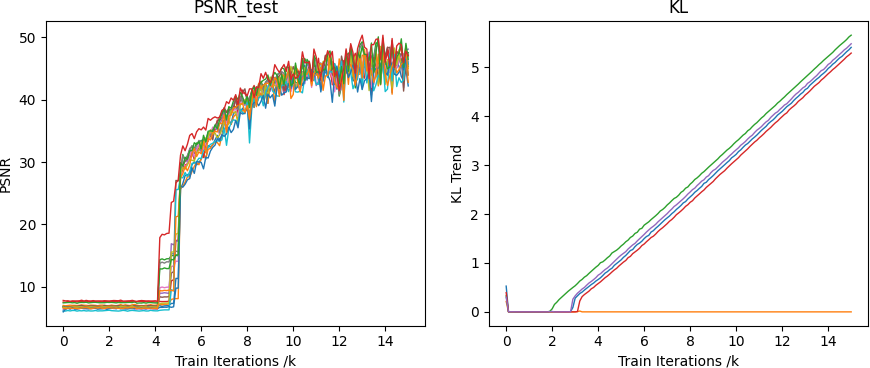}
  \caption{The convergence of beta-VAE with 5 latent variables, only 4 are activated.}
  \label{fig:1}
\end{figure}

\begin{figure}
  \centering
  \includegraphics[width=0.5\textwidth]{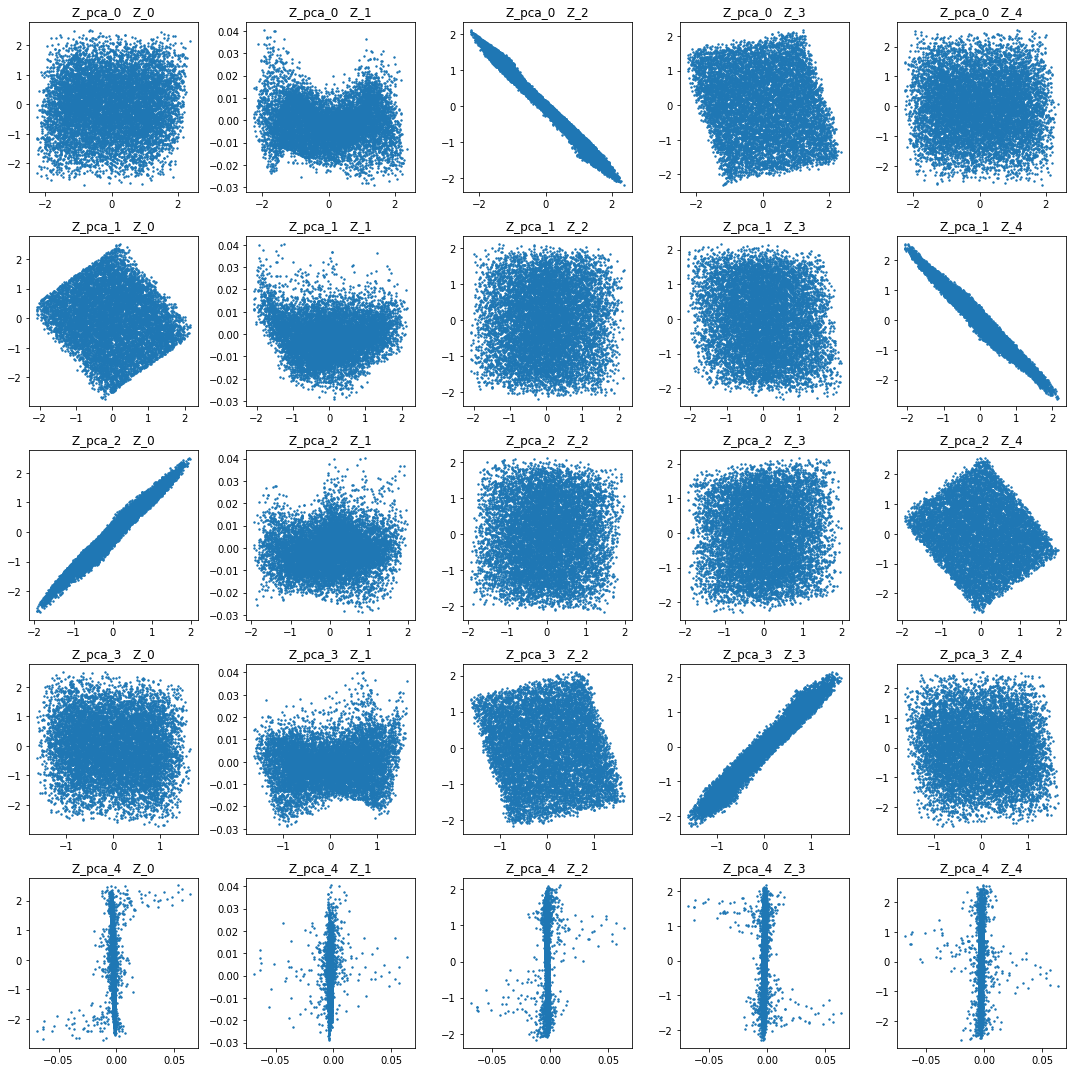}
  \caption{The components of PCA (5 dimensional) $Z_{pca0}$ to $Z_{pca4}$ with respect to the 5 latent variables learnt by beta-VAE $Z_0$ to $Z_4$.}
  \label{fig:4}
\end{figure}

We train the model given only 5 latent variables, slightly larger than that of the input dimension of 4. The convergence process is as shown in Fig.\ref{fig:1}. The 5 learnt variables are compared with the results of PCA, as shown in \ref{fig:4}, all the activated latent variables can find their corresponding PCA component. We can notice that the latent variables show the pattern of PCA, which learnt the most important/significant representation.

\subsubsection{Many Latent Variables Scenario}

\begin{figure}
  \centering
  \includegraphics[width=0.6\textwidth]{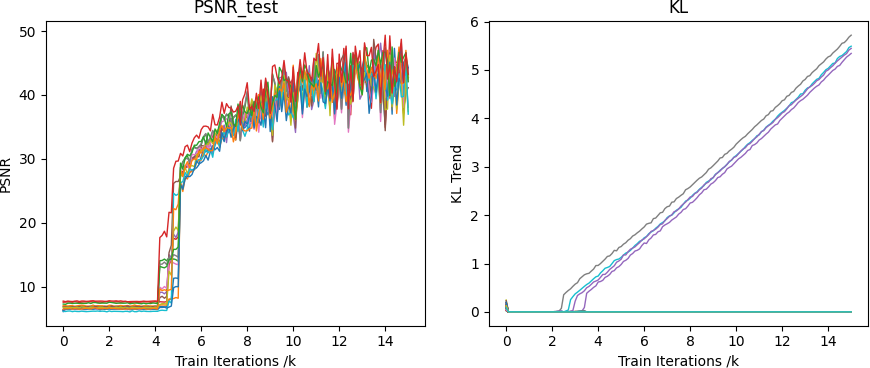}
  \caption{The convergence of beta-VAE with 100 latent variables, again, only 4 are activated.}
  \label{fig:5}
\end{figure}

\begin{figure}
  \centering
  \includegraphics[width=0.7\textwidth]{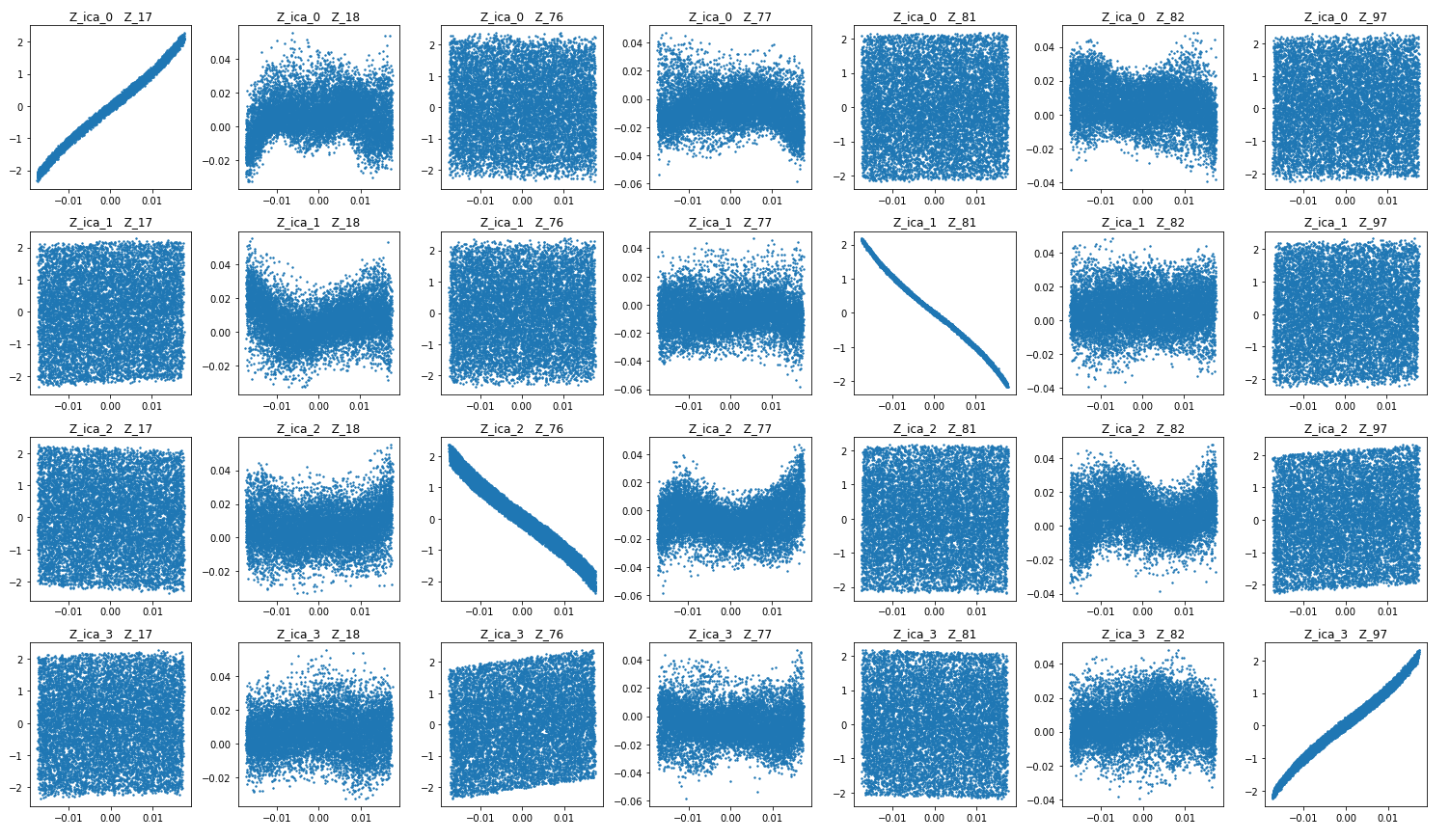}
  \caption{The components of ICA (4 dimensional) $Z_{ica0}$ to $Z_{ica4}$ with respect to the 100 latent variables learnt by beta-VAE $Z_0$ to $Z_100$. Due to the size limitations of the figure, we cropped out the activated and some of the unrelated latent variables, it can be observed that latent variable No.17, 76, 81 and 97 are corresponding to ICA components.}
  \label{fig:6}
\end{figure}

We train the model given 100 latent variables, much larger than that of the input dimension of 4. The convergence process is as shown in Fig.\ref{fig:5}. The 100 learnt variables are compared with the results of ICA, as shown in \ref{fig:6}, all the activated latent variables can find their corresponding ICA component. We can notice that the latent variables show the pattern of ICA, which learnt the disentangled representation.

\subsection{Non-Linear Dataset}

The non-linear data is created with an untrained 3-layer neural network, with input dimension of 4 and output dimension of 14. For each layer there are 14 neurons, activated with tanh activation function. The non-linear data is continuous and differentiable everywhere.

\subsubsection{PCA and ICA Results}

\begin{figure}
  \centering
  \includegraphics[width=0.4\textwidth]{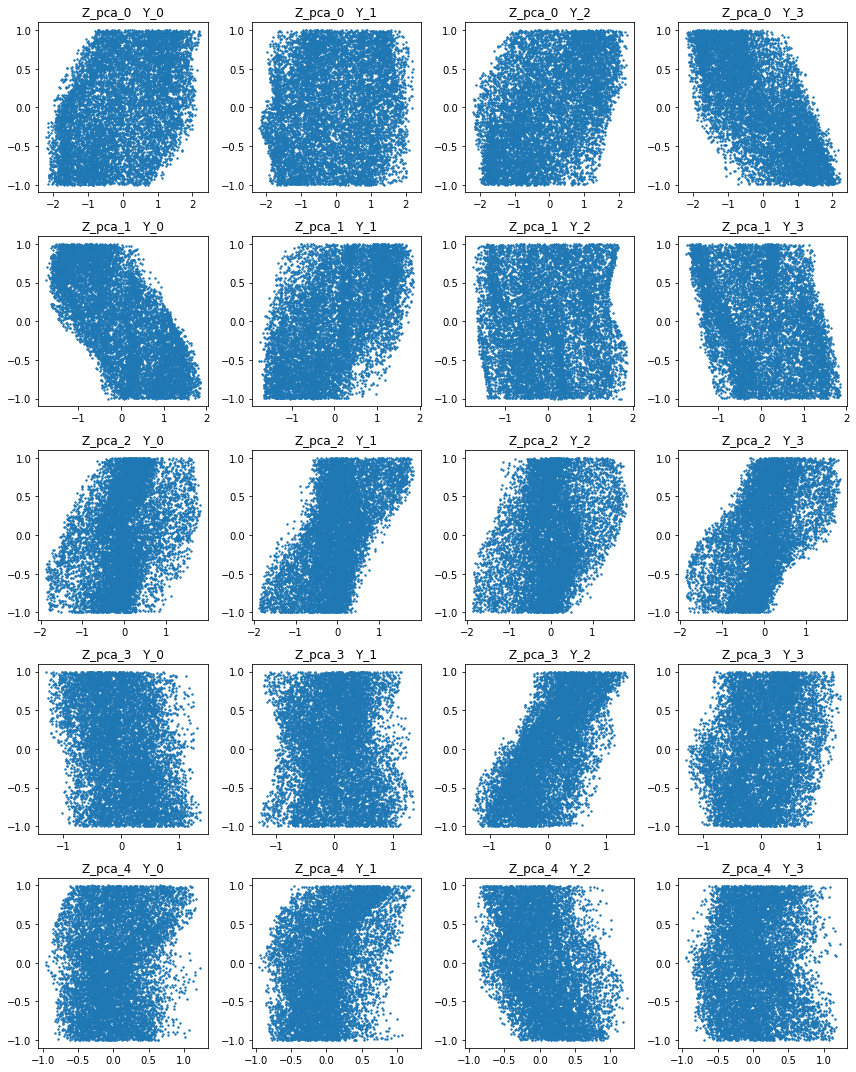}
  \caption{The components of PCA (5 dimensional) $Z_{pca0}$ to $Z_{pca4}$ with respect to the 4 input variables $Y_0$ to $Y_3$. No apparent pattern can be observed.}
  \label{fig:7}
\end{figure}

\begin{figure}
  \centering
  \includegraphics[width=0.4\textwidth]{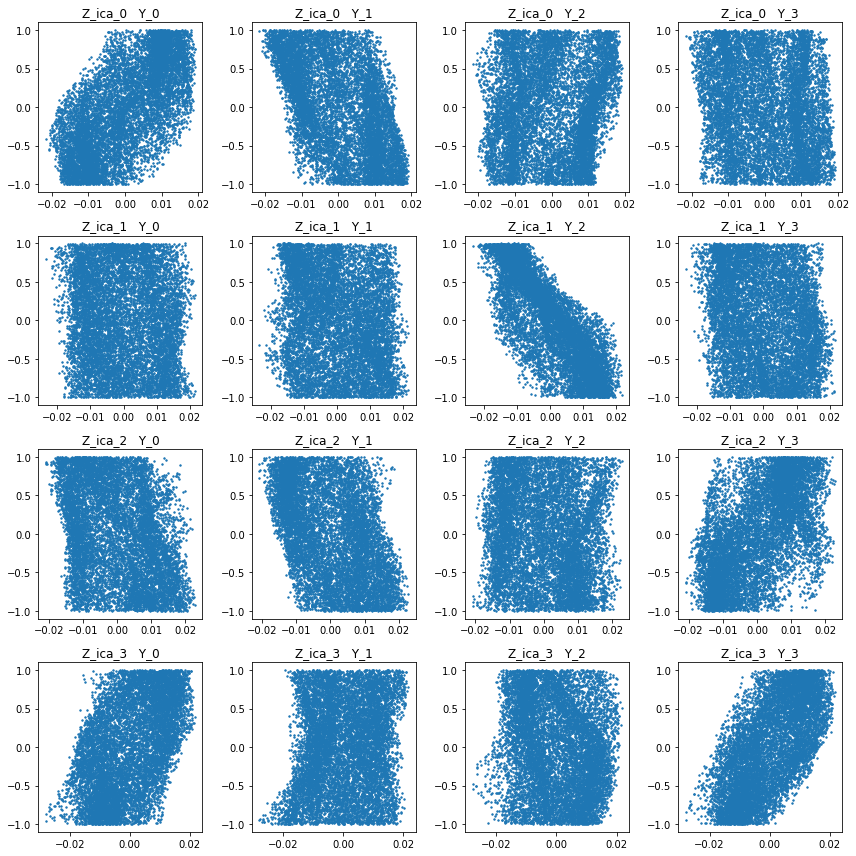}
  \caption{The components of ICA (4 dimensional) $Z_{ica0}$ to $Z_{ica4}$ with respect to the 4 input variables $Y_0$ to $Y_3$. The disentanglement is pretty obvious. No apparent pattern can be observed.}
  \label{fig:8}
\end{figure}

In the case of non-linear data, both PCA and ICA fail to extract anything useful, as shown in Fig.\ref{fig:7} and \ref{fig:8}

\subsubsection{Beta-VAE Results}

\begin{figure}
  \centering
  \includegraphics[width=0.6\textwidth]{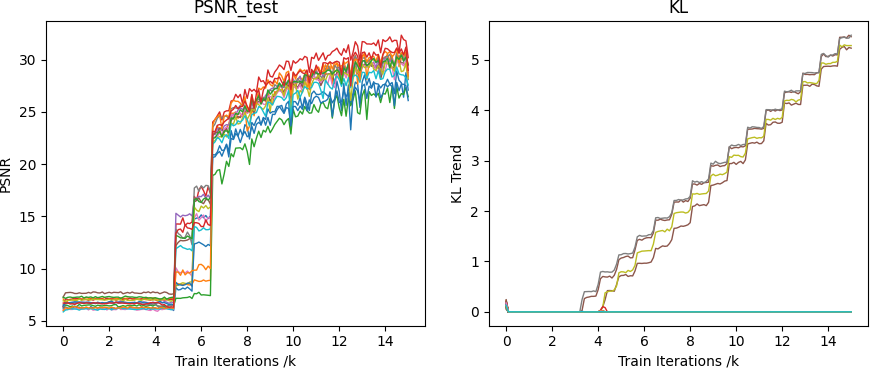}
  \caption{The convergence of beta-VAE with 5 latent variables on non-linear data, with 4 latent variables activated.}
  \label{fig:9}
\end{figure}

\begin{figure}
  \centering
  \includegraphics[width=0.6\textwidth]{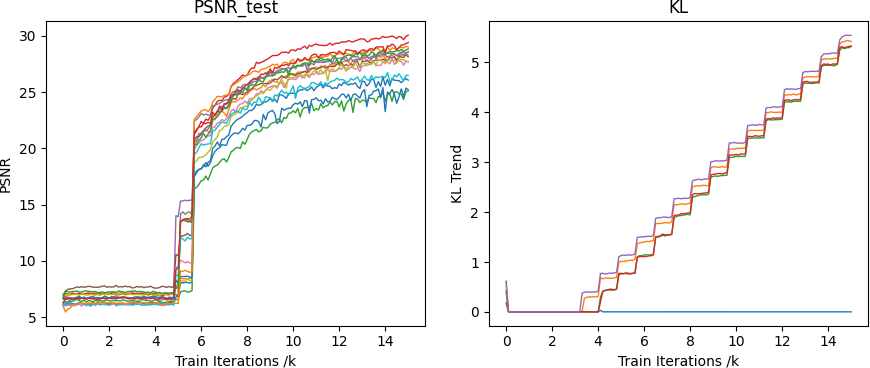}
  \caption{The convergence of beta-VAE with 100 latent variables on non-linear data, also with 4 latent variables activated.}
  \label{fig:10}
\end{figure}

We then conducted the experiments on beta-VAE with 5 and 100 latent variables. Due to the difficulties of non-linear data, we increased the gap between each $\beta$ shrink to 200. Both model converged with only 4 latent variables activated, as shown in Fig.\ref{fig:9} and \ref{fig:10}.

\begin{figure}
  \centering
  \includegraphics[width=0.4\textwidth]{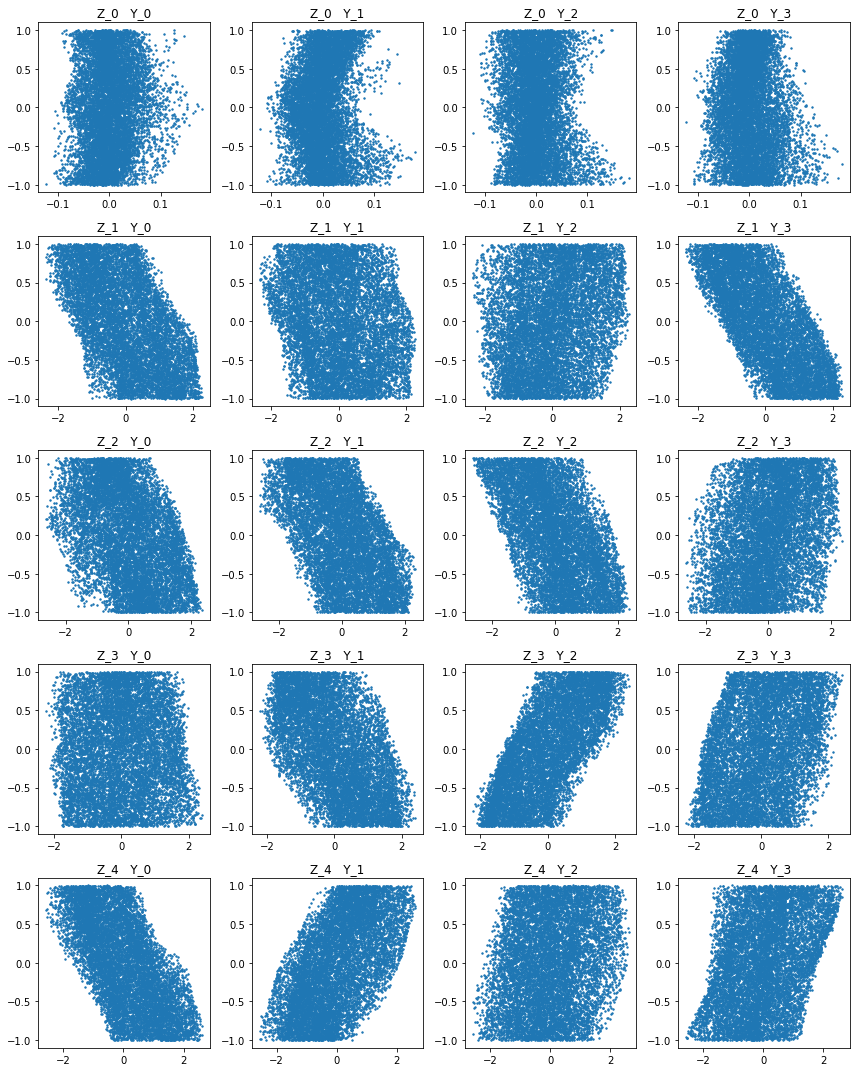}
  \caption{The latent variables of beta-VAE with 5 total latent variables $Z_{0}$ to $Z_{4}$ with respect to the 4 input variables $Y_0$ to $Y_3$.}
  \label{fig:11}
\end{figure}

\begin{figure}
  \centering
  \includegraphics[width=0.4\textwidth]{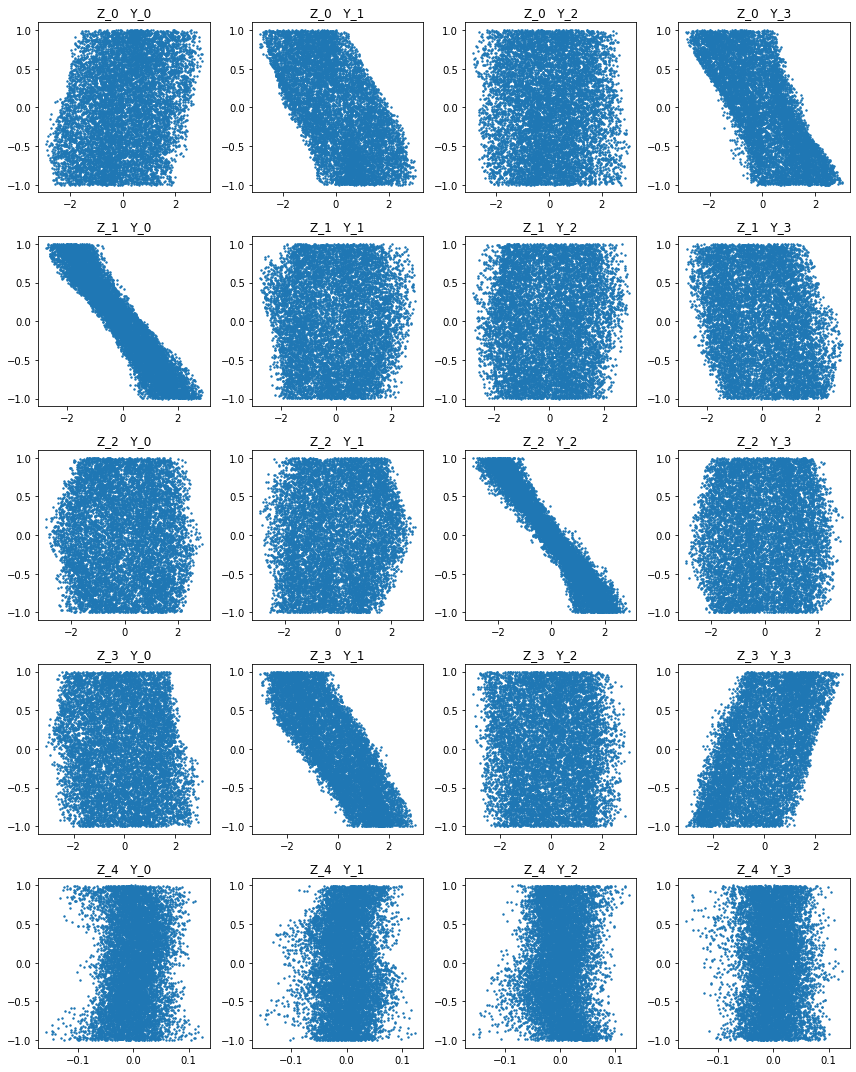}
  \caption{The selected 5 latent variables (4 of which are activated) of beta-VAE with 100 total latent variables $Z_{0}$ to $Z_{4}$ with respect to the 4 input variables $Y_0$ to $Y_3$. The selection is made by hand for better demonstration.}
  \label{fig:12}
\end{figure}

We then compared the learnt representation by the beta-VAE and the original input of the data, as shown in Fig.\ref{fig:11} and \ref{fig:12}. From the 2 figures of beta-VAE with 5 and 100 latent variables we can see that even though both model has untangled the non-linearity of the data, the one with 100 latent variables acquired representations closer to the original input. Apart from that, the model with 100 latent variables also converged to a better reconstruction result with testing set PSNR of 29.521 comparing to 27.790 of the other model. Thus proving that beta-VAE with 100 latent variables is better than that of 5 latent variables on non-linear data in terms of disentanglement of representation learnt and reconstruction quality.

\begin{figure}
  \centering
  \includegraphics[width=0.4\textwidth]{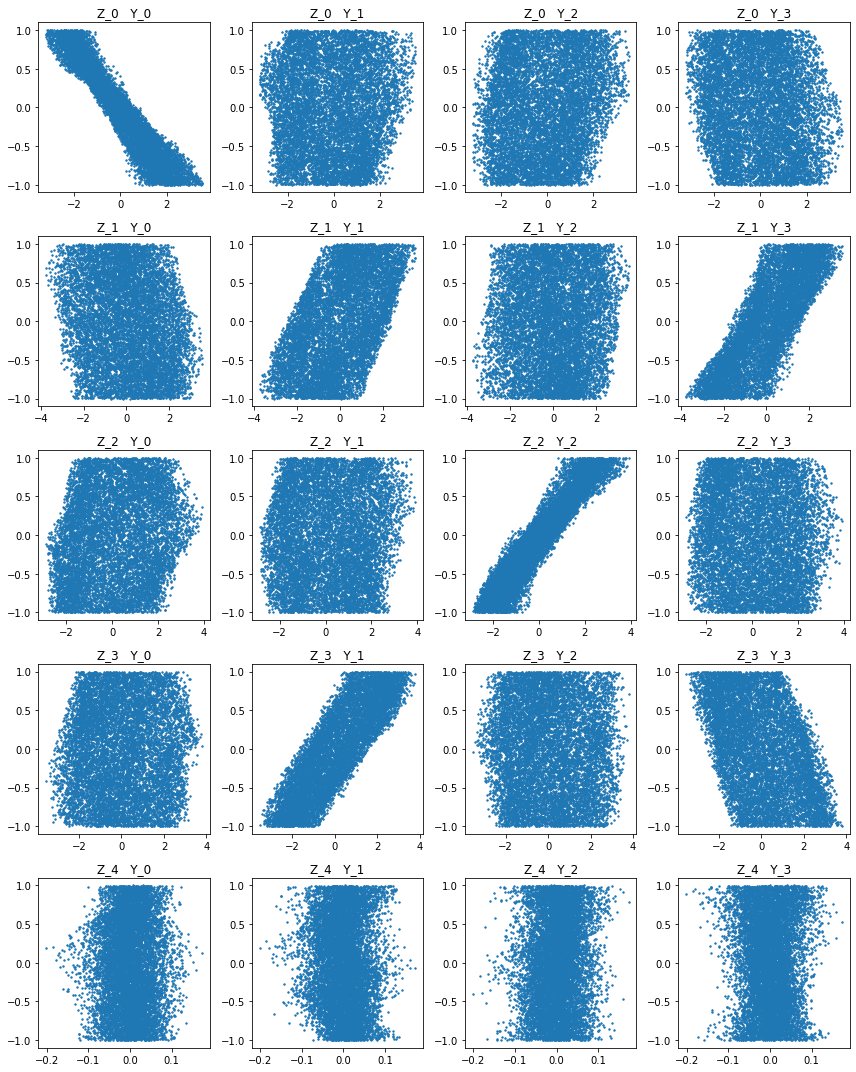}
  \caption{The selected 5 latent variables (4 of which are activated) of beta-VAE with 500 (wow) total latent variables $Z_{0}$ to $Z_{4}$ with respect to the 4 input variables $Y_0$ to $Y_3$. The selection is made by hand for better demonstration.}
  \label{fig:13}
\end{figure}

Last but not least, we conducted a final experiment this time the model have 500 latent variables, again, only 4 latent variables were activated. Fig.\ref{fig:13} shows the result. Comparing to Fig.\ref{fig:12}, it can be seen that the representation is even closer to that of the original input, and the reconstruction result is also higher, achieving 29.380.

\section{Discussion}

The experiments tell us that if we give beta-VAE more latent variables (can even exceed the data dimension itself, i.e. 100 is much bigger than 14), the representations learnt can be more disentangled, and if you are trying to find the most important variables, and do a non-linear PCA, you should use fewer latent variables.

We also proved again that beta-VAE is indeed a great model for representation learning, that very timidly, won't activate the latent variables when shouldn't.

While the exact reason for how the number of latent variables affect representation remain unknown, we come up with a few hypothesises:
1. More latent variables simply mean more candidates, and some candidates might be closer to that of the global optimal. To counter this we did try to eliminate the differences between latent variables by first suppressing all data that can come through the bottleneck of each latent variable to 0, meaning they are roughly on the same starting point.
2. The competition between the latent variables is what's causing this: while with fewer variables, the environment isn't very competitive, each latent variable activated only need to maximize the their own information that go through the bottleneck, which from a bigger picture, can harm the overall reconstruction results; in contrast, while there are more variables, they need to learn to adapt and form some kind of alliance, so the variables achieve a better overall result (learning a pattern more similar to the actual system input), resulting in getting more disentangled representations and better reconstruction results.

We also believe that once a representation is learnt, it will remain that way and further training won't affect it.

\bibliographystyle{elsarticle-harv} 

\end{document}